\documentclass[runningheads]{llncs}
\usepackage{graphicx}
\usepackage{comment}
\usepackage{amsmath,amssymb} % define this before the line numbering.
\usepackage{color}
\usepackage{graphicx}
\usepackage{wrapfig}
\usepackage{animate}
\usepackage[colorlinks=true]{hyperref}

\usepackage[sort,compress,square,comma,numbers]{natbib}
\PassOptionsToPackage{numbers, compress}{natbib}
\makeatletter
\renewcommand\bibsection%
{
  \section*{\refname
    \@mkboth{\MakeUppercase{\refname}}{\MakeUppercase{\refname}}}
}
\makeatother

\begin{document}
\pagestyle{headings}
\mainmatter
\def\ECCVSubNumber{4423} 
\newcommand{\etal}{\textit{et al}. }
\newcommand{\ie}{\textit{i}.\textit{e}. }
\newcommand{\eg}{\textit{e}.\textit{g}. }

\title{It is not the Journey but the Destination: \\
Endpoint Conditioned Trajectory Prediction}

\begin{comment}
\titlerunning{ECCV-20 submission ID \ECCVSubNumber} 
\authorrunning{ECCV-20 submission ID \ECCVSubNumber} 
\author{Anonymous ECCV submission}
\institute{Paper ID \ECCVSubNumber}
\end{comment}
%******************
\titlerunning{PECNet: Pedestrian Endpoint Conditioned Trajectory Prediction Network}

\author{Karttikeya Mangalam\inst{1}, % \and
Harshayu Girase\inst{1}, % \and
Shreyas Agarwal\inst{1}, % \and
Kuan-Hui Lee\inst{2}, 
Ehsan Adeli\inst{3},  
Jitendra Malik\inst{1}, Adrien Gaidon\inst{2}}
\authorrunning{K. Mangalam, H. Girase, S. Agarwal, K. Lee, E. Adeli, J. Malik, A. Gaidon}

\institute{University of California, Berkeley \and
Toyota Research Institute 
\and
Stanford University\\
\email{mangalam@cs.berkeley.edu}}
% \end{comment}
\newcommand{\KM}[1]{\textcolor{blue}{\textbf{KM:} #1}}
%******************
\maketitle
\begin{abstract}
Human trajectory forecasting with multiple socially interacting agents is of critical importance for autonomous navigation in human environments, e.g., for self-driving cars and social robots. In this work, we present Predicted Endpoint Conditioned Network (PECNet) for flexible human trajectory prediction. PECNet infers distant trajectory endpoints to assist in long-range multi-modal trajectory prediction. A novel non-local social pooling layer enables PECNet to infer diverse yet socially compliant trajectories. Additionally, we present a simple ``truncation-trick" for improving diversity and multi-modal trajectory prediction performance. We show that PECNet improves state-of-the-art performance on the Stanford Drone trajectory prediction benchmark by $\sim 20.9\%$ and on the ETH/UCY benchmark by $\sim40.8\%$\footnote{ Code available at project homepage: \href{https://karttikeya.github.io/publication/htf/}{https://karttikeya.github.io/publication/htf/}}.
\end{abstract}
\begin{figure}
    \centering
    \includegraphics[width=0.9\textwidth]{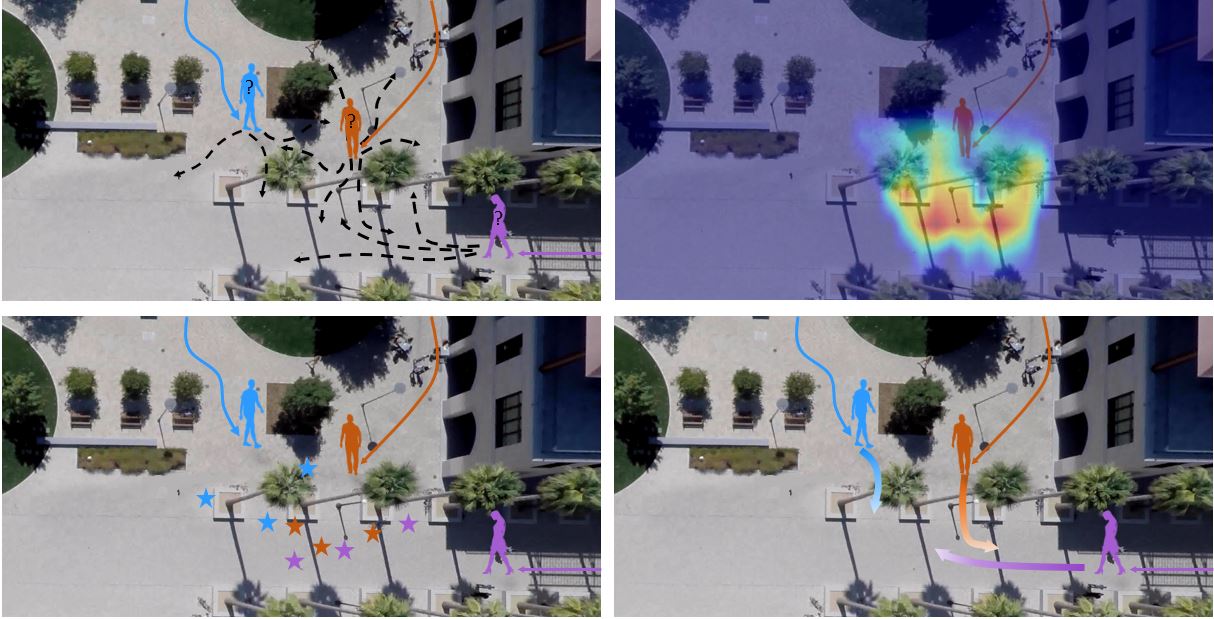} \\
    \caption{\textbf{Imitating the Human Path Planning Process}. Our proposed approach to model pedestrian trajectory prediction (top left) breaks down the task in two steps: (a) inferring the local endpoint distribution (top right), and then (b) conditioning on sampled future endpoints (bottom left) for jointly planning socially compliant trajectories for all the agents in the scene (bottom right).}
    \label{pull_fig}
\end{figure}
\section{Introduction}
\label{sec:introduction}
Predicting the movement of dynamic objects is a central problem for autonomous agents, be it humans, social robots \cite{bennewitz2002learning}, or self-driving cars \cite{thrun2005probabilistic}. Anticipation by prediction is indeed required for smooth and safe path planning in a changing environment.
One of the most frequently encountered dynamic objects are humans. Hence, predicting human motion is of paramount importance for navigation, planning, human-robot interaction, and other critical robotic tasks. However, predicting human motion is nuanced, because humans are not inanimate entities evolving under Newtonian laws \cite{baker2009action}. Rather, humans have the will to exert causal forces to change their motion and constantly adjust their paths as they navigate around obstacles to achieve their goals \cite{ziebart2009planning}. This complicated planning process is partially internal, and thus makes predicting human trajectories from observations challenging. Hence, a multitude of aspects should be taken into account beyond just past movement history, for instance latent predetermined goals, other moving agents in the scene, and social behavioral patterns.

In this work, we propose to address \textbf{human trajectory prediction by modeling intermediate stochastic goals we call endpoints}. We hypothesize that three separate factors interact to shape the trajectory of a pedestrian. First, we posit that pedestrians have some understanding of their long-term desired destination. We extend this hypothesis to sub-trajectories, i.e.\ the pedestrian has one or multiple intermediate destinations, which we define as potential endpoints of the local trajectory. These sub-goals can be more easily correlated with past observations to predict likely next steps and disentangle potential future trajectories.

Second, the pedestrian plans a trajectory to reach one of these sub-goals, taking into account the present scene elements. Finally, as the agent go about executing a plan, the trajectory gets modified to account for other moving agents, respecting social norms of interaction.

Following the aforementioned intuition, we propose to decompose the trajectory prediction problem into two sub-problems that also motivate our proposed architecture (Figure \ref{pull_fig}). First, given the previous trajectories of the humans in the scene, we propose to estimate a latent belief distribution modeling the pedestrians' possible endpoints. Using this estimated latent distribution, we sample plausible endpoints for each pedestrian based on their observed trajectory. A socially-compliant future trajectory is then predicted, conditioned not only on the pedestrian and their immediate neighbors' histories (observed trajectories) but also everybody's estimated endpoints.

In conclusion, our contribution in this work is threefold. \textbf{First}, we propose a socially compliant, endpoint conditioned variational auto-encoder that closely imitates the multi-modal human trajectory planning process. \textbf{Second}, we propose a novel self-attention based social pooling layer that generalizes previously proposed social pooling mechanisms. \textbf{Third}, we show that our model can predict stable and plausible intermediate goals that enable setting a new state-of-the-art on several trajectory prediction benchmarks, improving by \textbf{20.9}\% on SDD \cite{robicquet2016learning} \& \textbf{40.8\%} on ETH \cite{pellegrini2010improving} \& UCY \cite{lerner2007crowds}.

\section{Related work}

There have been many previous studies \cite{2019pedtrajpredsurvey} on how to forecast pedestrians' trajectories and predict their behaviors.
Several previous works propose to learn statistical behavioral patterns from the observed motion trajectories \cite{kruse1998camera,liao2003voronoi,bennewitz2005learning,tay2008modelling,kafer2010recognition,aoude2011mobile,keller2013will,goldhammer2014pedestrian,xiao2015unsupervised,kucner2017enabling} for future trajectory prediction.
Since then, many studies have developed models to account for agent interactions that may affect the trajectory --- specifically, through scene and/or social information. Recently, there has been a significant focus on multi-modal trajectory prediction to capture different possible future trajectories given the past. There has also been some research on goal-directed path planning, which consider pedestrians' goals while predicting a path.

\subsection{Context-Based Prediction}
Many previous studies have imported environment semantics, such as crosswalks, road, or traffic lights, to their proposed trajectory prediction scheme.
Kitani et al. \cite{kitani2012activity} encode agent-space interactions by a Markov Decision Process (MDP) to predict potential trajectories for an agent.
Ballan et al. \cite{ballan2016knowledge} leverage a dynamic Bayesian network to construct motion dependencies and patterns from training data and transferred the trained knowledge to testing data.
With the great success of the deep neural network, the Recurrent Neural Network (RNN) has become a popular modeling approach for sequence learning.
Kim et al. \cite{kim2017probabilistic} train a RNN combining multiple Long Short-term Memory (LSTM) units to predict the location of nearby cars. 
These approaches incorporate rich environment cues from the RGB image of the scene for pedestrians' trajectory forecasting.

 Behaviour of surrounding dynamic agents is also a crucial cue for contextual trajectory prediction. Human behavior modeling studied from a crowd perspective, \ie, how a pedestrian interacts with other pedestrians, has also been studied widely in human trajectory prediction literature.
Traditional approaches use \textit{social forces} \cite{helbing1995social,mehran2009abnormal,yamaguchi2011you,alahi2014socially} to capture pedestrians' trajectories towards their goals with attractive forces, while avoiding collisions in the path with repulsive forces. 
These approaches require hand-crafted rules and features, which are usually complicated and insufficiently robust for complicated high-level behavior modeling.
Recently, many studies applied Long Short Term Memory (LSTM \cite{hochreiter1997long}) networks to model trajectory prediction with the social cues.
Alahi et al. \cite{alahi2016social} propose a Social LSTM which learns to predict a trajectory with joint interactions.
Each pedestrian is modeled by an individual LSTM, and LSTMs are connected with their nearby individual LSTMs to share information from the hidden state.

\subsection{Multimodal Trajectory Prediction}
In \cite{lee2017desire,gupta2018social}, the authors raise the importance of accounting for the inherent multi-modal nature of human paths  \ie, given pedestrians' past history, there are many plausible future paths they can take. This shift of emphasis to plan for multiple future paths has led many recent works to incorporate multi-modality in their trajectory prediction models. 
 Lee et al. \cite{lee2017desire} propose a conditional variational autoencoder (CVAE), named DESIRE, to generate multiple future trajectories based on agent interactions, scene semantics and expected reward function, within a sampling-based inverse optimal control (IOC) scheme. 
In \cite{gupta2018social}, Gupta et al. propose a Generative Adversarial Network (GAN) \cite{goodfellow2014generative} based framework with a novel social pooling mechanism to generate multiple future trajectories in accordance to social norms. In \cite{sadeghian2019sophie}, Sadeghian et al. also propose a GAN based framework named SoPhie, which utilizes path history of all the agents in a scene and the scene context information. SoPhie employs a social attention mechanism with physical attention, which helps in learning social information across the agent interactions.
However, these socially-aware approaches do not take into account the pedestrians' ultimate goals, which play a key role in shaping their movement in the scene. 
A few works also approach trajectory prediction via an inverse reinforcement learning (IRL) setup. Zou et al. \cite{zou2018understanding} applies Generative Adversarial Imitation Learning (GAIL) \cite{ho2016generative} for trajectory prediction, named Social-Aware GAIL (SA-GAIL).
With IRL, the authors model the human decision-making process more closely through modeling humans as agents with states (past trajectory history) and actions (future position). SA-GAIL generates socially acceptable trajectories via a learned reward function.

\subsection{Conditioned-on-Goal}
Goal-conditioned approaches are regarded as inverse planning or \textit{prediction by planning} where the approach learns the final intent or goal of the agent before predicting the full trajectory.
In \cite{rehder2015goal}, Rehder \etal propose a particle filtering based approach for modeling destination conditioned trajectory prediction and use explicit Von-Mises distribution based probabilistic framework for prediction. 
Later in a follow-up work, \cite{rehder2018pedestrian} Rehder \etal further propose a deep learning based destination estimation approach to tackle intention recognition and trajectory prediction simultaneously.
The approach uses fully Convolutional Neural Networks (CNN) to construct the path planning towards some potential destinations which are provided by a recurrent Mixture Density Network (RMDN). While both the approaches make an attempt for destination conditioned prediction, a fully probabilistic approach trains poorly due to unstable training and updates. Further, they ignore the presence of other pedestrians in the scene which is key for predicting shorter term motions which are missed by just considering the environment.
Rhinehart et al. \cite{rhinehart2019precog} propose a goal-conditioned multi-agent forecasting approach named PRECOG, which learns a probabilistic forecasting model conditioned on drivers' actions intents such as ahead, stop, etc. However, their approach is designed for vehicle trajectory prediction, and thus conditions on semantic goal states. In our work, we instead propose to utilize destination position for pedestrian trajectory prediction. 

 In \cite{li2019conditional}, Li \etal posit a Conditional Generative Neural System (CGNS), the previous established state-of-the-art result on the ETH/UCY dataset. They propose to use variational divergence minimization with soft-attention to predict feasible  multi-modal trajectory distributions. 
Even more recently, Bhattacharyya \etal \cite{bhattacharyya2019conditional} propose a conditional flow VAE that proposed a general normalizing flow for structured sequence prediction and applies it to the problem of trajectory prediction. Concurrent to our work, Deo \etal \cite{deo2020trajectory} propose P2TIRL, a Maximum Entropy Reinforcement Learning based trajectory prediction module over a discrete grid. The work  \cite{bhattacharyya2019conditional} shares state-of-the-art with \cite{deo2020trajectory} on the Stanford Drone Dataset (SDD) with the TrajNet \cite{sadeghiankosaraju2018trajnet} split. However, these works fail to consider the human aspect of the problem, such as interaction with other agents. We compare our proposed PECNet with all three of the above works on both the SDD \& ETH/UCY datasets.  

\section{Proposed Method}
In this work, we aim to tackle the task of human trajectory prediction by reasoning about all the humans in the scene jointly while also respecting social norms. Suppose a pedestrian $p^k$ enters a scene $\mathcal{I}$. Given the previous trajectory of $p$ for past $t_p$ steps, as a sequence of coordinates $\mathcal{T}^k_p := \{\textbf{u}^k\}_{i=1}^{t_p}  =\{(x^k,y^k)\}_{i=1}^{t_p}$, the problem requires predicting the future position of $p^k$ on $\mathcal{I}$ for next $t_f$ steps, $\mathcal{T}^k_f := \{\textbf{u}^k\}_{i=t_p + 1}^{t_p + t_f} = \{(x,y)\}_{i=t_p + 1}^{t_p + t_f}$. 

As mentioned in Section \ref{sec:introduction}, we break the problem into two daisy chained steps. First, we model the sub-goal of $p^k$, i.e.\ the last observed trajectory points of $p^k$ say, $\mathcal{G}^k = \textbf{u}^k|_{t_p + t_f}$ as a representation of the predilection of $p^k$ to go its pre-determined route. This sub-goal, also referred to as the endpoint of the trajectory, the pedestrian's desired end destination for the current sequence. Then in the second step, we jointly consider the past histories $\{\mathcal{T}^k_p\}_{k=1}^\alpha$ of all the pedestrians $\{p^{k}\}_{k=1}^{\alpha}$ present in the scene and their estimated endpoints $\{\mathcal{G}^k\}_{k=1}^\alpha$ for predicting socially compliant future trajectories $\mathcal{T}_f^k$. In the rest of this section we describe in detail, our approach to achieve this, using the endpoint estimation VAE for sampling the future endpoints $\mathcal{G}$ and a trajectory prediction module to use the sampled endpoints $\hat{\mathcal{G}}^k$ to predict $\mathcal{T}_f$. 
\begin{figure}[t!]
    \centering
    \includegraphics[width=\textwidth]{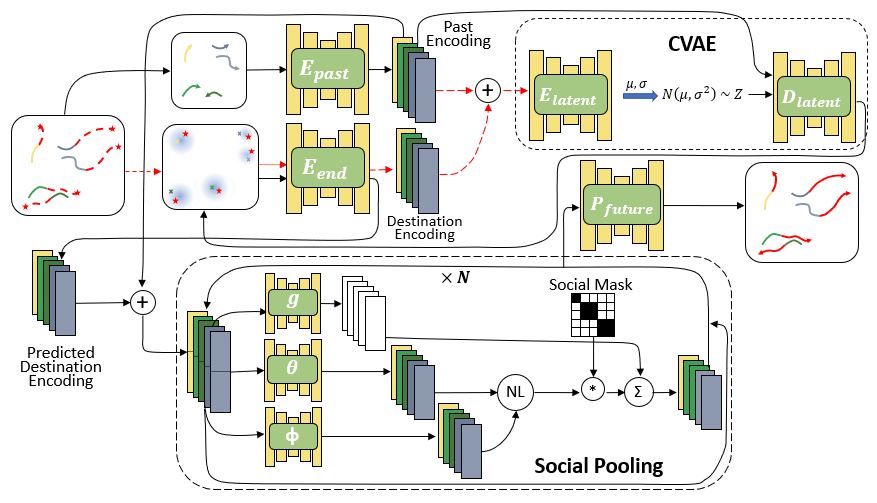} \\
    \caption{\textbf{Architecture of PECNet}: PECNet uses past history, $\mathcal{T}_i$ along with ground truth endpoint $\mathcal{G}_c$ to train a VAE for multi-modal endpoint inference. Ground-truth endpoints are denoted by $\star$ whereas \textbf{x} denote the sampled endpoints $\hat{\mathcal{G}}_c$. The sampled endpoints condition the social-pooling \& predictor networks for multi-agent multi-modal trajectory forecasting. Red connections denote the parts utilized only during training. Shades of the same color denote spatio-temporal neighbours encoded with the block diagonal social mask in social pooling module. Further Details in Section \ref{sec:vae}.}
    \label{fig:arch}
\end{figure}

\subsection{{Endpoint} VAE}
\label{sec:vae}
We propose to model the predilection of the pedestrian as a sub-goal endpoint $\mathcal{G} := \textbf{u}_{t_f} = (x_{t_f}, y_{t_f})$ which is the last observed trajectory point for pedestrian $p^k$. First, we infer a distribution on $\mathcal{G}$ based on the previous location history $\mathcal{T}_i$ of $p^k$ using the Endpoint VAE. 

As illustrated in Figure \ref{fig:arch}, we extract the previous history $\mathcal{T}_i^k$ and the ground truth endpoint $\mathcal{G}^k$ for all pedestrian $p_k$ in the scene. We encode the past trajectory  $\mathcal{T}_i^k$ of all $p_k$ independently using a past trajectory encoder $\mathbf{E}_{\textit{past}}$. This yields us $\mathbf{E}_{\textit{past}}(\mathcal{T}_i)$, a representation of the motion history. Similarly, the future endpoint  $\mathcal{G}^k$ is encoded with an Endpoint encoder $\mathbf{E}_{\textit{end}}$ to produce $\mathbf{E}_{\textit{end}}(\mathcal{G}^k)$ independently for all $k$. These representations are concatenated together and passed into the latent encoder $\mathbf{E}_{\textit{latent}}$ which produces parameter $(\boldsymbol{\mu},\boldsymbol{\sigma})$ for encoding the latent variable $z = \mathcal{N}(\boldsymbol{\mu},\boldsymbol{\sigma})$ of the VAE. Finally, we sample possible latent future endpoints from $\mathcal{N}(\boldsymbol{\mu},\boldsymbol{\sigma})$, concatenate it with $\mathbf{E}_{\textit{past}}(\mathcal{T}_i)$ for past context and decode using the latent decoder $\mathbf{D}_{\textit{latent}}$ to yield our guesses for $\hat{\mathcal{G}}^k$. Since the ground truth ${\mathcal{G}}^k$ belongs to the future, and is unavailable at test time, during evaluation we sample $z$ from $\mathcal{N}(0,\sigma_T \mathbf{I})$, concatenate with $\mathbf{E}_{\textit{past}}(\mathcal{T}_i)$ (as done in training) and then use the learned $\mathbf{D}_{\textit{latent}}$ to estimate the future $\hat{\mathcal{G}}^k$. This is illustrated in Figure \ref{fig:arch} where the red connections are only used in the training and not in the evaluation phase. 

\noindent \textbf{Truncation trick}: In \cite{brock2018large}, Brock \etal introduce the `Truncation Trick' as a method of trade-off between the fidelity and variety of samples produced by the generator in BigGAN. In this work, we propose an analogous trick for evaluation phase in multi-modal trajectory forecasting where the variance of the latent endpoint sampling distribution is changed according to the number of samples ($K$) allowed for multi-modal prediction.  In a situation requiring few shot multi-modal prediction, such as under computation constraints, where only a few samples ($K = 1,2$ or $3$) are permissible, we propose to use $\sigma_T = 1$ and truncate the sampling distribution at $\pm c\sqrt{K-1}$. In contrast, in situations where a high number of predictions are to be generated (such as $K= 20$, a standard setting on benchmarks) we propose to use $\sigma_T > 1$ with no truncation. We posit that this procedure allows simple adjustment of prediction diversity in favor of overall performance for different $K$, thereby providing a simple method of achieving good performance in all settings without requiring any retraining.     

\subsection{Endpoint conditioned Trajectory Prediction}
\label{sec:method_joint_traj}
Using the sampled estimate of the \textit{endpoints} $\mathcal{\hat{G}}$ from Endpoint VAE, we employ the endpoint encoder $\mathbf{E}_{\textit{end}}$ once again (within the same forward pass) to obtain encodings for the sampled endpoints $\mathbf{E}_{\textit{end}}(\mathcal{\hat{G}}^k)$. This is used along with prediction network to \textit{plan} the path $\mathcal{T}_f$ starting to $\mathcal{G}$ thereby predicting the future path.

 Note that, another design choice could have been that even during training, the ground truth $\mathbf{E}_{\textit{end}}(\mathcal{{G}}^k)$ are used to predict the future $\mathcal{T}_f$. This seems reasonable as well since it provides cleaner, less noisy signals for the downstream social pooling \& prediction networks while still training the overall module end to end (because of coupling through $\mathbf{E}_{\textit{past}}$). However, such a choice will decouple training of the Endpoint VAE (which would then train only with KL Divergence and AWL loss, refer Section \ref{sec:Loss}) and social pooling network (which would then train only with ATL loss, refer \ref{sec:Loss}) leading to inferior performance empirically.  
 
 The sampled endpoints' representations $\mathbf{E}_{\textit{end}}(\mathcal{\hat{G}}^k)$ are then concatenated with corresponding $\mathbf{E}_{\textit{past}}(\mathcal{T}_i)$ (as in Section \ref{sec:vae}) and passed through $N$ rounds of social pooling using a social pooling mask $\mathbf{M}$ for all the pedestrians in the scene jointly. The social pooling mask $\mathbf{M}$ is $\alpha \times \alpha$ block diagonal matrix denoting the social neighbours for all $\{p_i\}_{i=1}^\alpha$ pedestrians in the scene. Mathematically,
\begin{equation}
    \mathbf{M}[i,j] =
    \begin{cases}
      0 & \text{if}\ \displaystyle\min_{1 \leq m,n \leq t_p} \Vert \mathbf{u}_m^i - \mathbf{u}_n^j \Vert_2 > t_{\textit{dist}} \\
    0 & \text{if} \displaystyle\min_{1 \leq m \leq t_p} \vert\mathcal{F}(\mathbf{u}_0^i) - \mathcal{F}(\mathbf{u}_m^j) \vert * \displaystyle\min_{1 \leq m \leq t_p} \vert\mathcal{F}(\mathbf{u}_m^i) - \mathcal{F}(\mathbf{u}_0^j) \vert) > 0\\
    %   0 & i = j  \\
      1 & \text{otherwise}
    \end{cases}
\end{equation}
where $\mathcal{F}(.)$ denoted the actual frame number the trajectory was observed at. Intuitively, $\mathbf{M}$ defines the spatio-temporal neighbours of each pedestrian $p_i$ using proximity threshold $t_{\textit{dist}}$ for distance in space and ensure temporal overlap. Thus, the matrix $\mathbf{M}$ encodes crucial information regarding social locality of different trajectories which gets utilized in attention based pooling as described below.

\noindent\textbf{Social Pooling}: Given the concatenated past history and sampled way-point representations $X^{(1)}_k = (\mathbf{E}_{\textit{past}}(\mathcal{T}^k_p),\mathbf{E}_{\textit{end}}(\mathcal{\hat{G}}^k))$ we do $N$ rounds of social pooling where the $(i+1)$th round of pooling recursively updates the representations $X^{(i)}_k$ from the last round according to the non-local attention mechanism \cite{wang2018non}: %given by,
\begin{equation}
X^{(i+1)}_k = X^{(i)}_k + \frac{1}{ \displaystyle\sum_{j = 1}^\alpha \mathbf{M}_{ij} \cdot e^{\boldsymbol{\phi}(X^{(i)}_k)^{T}\boldsymbol{\theta}(X^{(i)}_j)}}
\sum_{j = 1}^\alpha \mathbf{M}_{ij} \cdot e^{\boldsymbol{\phi}(X^{(i)}_k)^{T}\boldsymbol{\theta}(X^{(i)}_j)}\mathbf{g}(X^{(i)}_k)
\end{equation}
where $\{\boldsymbol{\theta}, \boldsymbol{\phi}\}$ are encoders of $X_k$ to map to a learnt latent space where the representation similarity between $p_i$ and $p_j$ trajectories is calculated using the embedded gaussian $\exp ({\boldsymbol{\phi}(X_k)^{T}\boldsymbol{\theta}(X_j)})$ for each round of pooling. The social mask, $\mathbf{M}$ is used point-wise to allow pooling only on the spatio-temporal neighbours masking away other pedestrians in the scene. Finally, $\boldsymbol{g}$ is a transformation encoder for $X_k$ used for the weighted sum with all other neighbours. The whole procedure, after being repeated $N$ times yields $X_k^{(N)}$, the pooled prediction features for each pedestrian with information about the past positions and future destinations of all other neighbours in the scene. 

 Our proposed social pooling is a novel method for extracting relevant information from the neighbours using non-local attention. The proposed social non local pooling (S-NL) method is permutation \textit{invariant} to pedestrian indices as a useful inductive bias for tackling the social pooling task. Further, we argue that this method of learnt social pooling is more robust to social neighbour mis-identification such as say, mis-specified distance ($t_\textit{dist}$) threshold compared to previously proposed method such as max-pooling \cite{gupta2018social}, sorting based pooling \cite{sadeghian2019sophie} or rigid grid-based pooling \cite{alahi2016social} since a learning based method can ignore spurious signals in the social mask $\mathbf{M}$. 

 The pooled features $X_k^{(N)}$ are then passed through the prediction network $\mathbf{P}_\textit{future}$ to yield our estimate of rest of trajectory $\{\mathbf{u}^k\}_{k = t_p + 1}^{t_p + t_f}$ which are concatenated with sampled endpoint $\hat{\mathcal{G}}$ yields $\hat{\mathcal{T}}_f$. %IMFIX
The complete network is trained end to end with the losses described in the next subsection.
\subsection{Loss Functions}
\label{sec:Loss}
For training the entire network end to end we use the loss function, 
\begin{equation}
\mathcal{L}_{PECNet} = \lambda_1 \underbrace{D_{KL}(\mathcal{N}(\boldsymbol{\mu}, \boldsymbol{\sigma}) \Vert \mathcal{N}(0, \mathbf{I}))}_{\textit{KL Div in latent space}} + \lambda_2 \underbrace{\Vert\hat{\mathcal{G}}_c - \mathcal{G}_c\Vert^2_2}_{AEL} +  \underbrace{\Vert \hat{\mathcal{T}}_f - \mathcal{T}_f \Vert^2}_{ATL}
\end{equation}
 % IMFIX
where the KL divergence term is used for training the Variational Autoencoder, the Average endpoint Loss (AEL) trains $\mathbf{E}_{\textit{end}}$, $\mathbf{E}_{\textit{past}}$, $\mathbf{E}_{\textit{latent}}$ and $\mathbf{D}_{\textit{latent}}$ and the Average Trajectory Loss (ATL) trains the entire module together.    

\section{Experiments}
\subsection{Datasets}

\noindent\textbf{Stanford Drone Dataset}: Stanford Drone Dataset \cite{robicquet2016learning} is a well established benchmark for human trajectory prediction in bird's eye view. The dataset consists of 20 scenes captured using a drone in top down view around the university campus containing several moving agents like humans and vehicles. It consists of over $11,000$ unique pedestrians capturing over $185,000$ interactions between agents and over $40,000$ interactions between the agent and scene \cite{robicquet2016learning}. We use the standard test train split as used in \cite{sadeghian2019sophie,gupta2018social,deo2020trajectory} and other previous works. 

\noindent\textbf{ETH/UCY}: Second is the ETH \cite{pellegrini2010improving} and UCY \cite{lerner2007crowds} dataset group, which consists of five different scenes -- ETH \& HOTEL (from ETH) and UNIV, ZARA1, \& ZARA2 (from UCY). All the scenes report the position of pedestrians in world-coordinates and hence the results we report are in metres. The scenes are captured in unconstrained environments with few objects blocking pedestrian paths. Hence, scene constraints from other physical non-animate entities is minimal. For bench-marking, we follow the commonly used leave one set out strategy i.e., training on four scenes and testing on the fifth scene \cite{sadeghian2019sophie,gupta2018social,li2019conditional}.

\setlength{\columnsep}{10pt}
\setlength{\intextsep}{5pt}
\begin{wrapfigure}{r}{0.5\textwidth}
\centering
\begin{tabular}{|c|c|}
\hline
& Network Architecture \\ \hline
$\mathbf{E}_{\textit{way}}$ & $2 \rightarrow 8 \rightarrow 16 \rightarrow 16$   \\
$\mathbf{E}_{\textit{past}}$ & $16 \rightarrow 512 \rightarrow 256 \rightarrow 16$ \\ \hline \hline
$\mathbf{E}_{\textit{latent}}$ & $32 \rightarrow 8 \rightarrow 50 \rightarrow 32$   \\ 
$\mathbf{D}_{\textit{latent}}$ & $32 \rightarrow 1024 \rightarrow 512 \rightarrow 1024 \rightarrow 2$   \\  \hline \hline
$\boldsymbol{\phi, \theta}$ & $32 \rightarrow 512 \rightarrow 64 \rightarrow 128$   \\ 
$\mathbf{g}$ & $32 \rightarrow 512 \rightarrow 64 \rightarrow 32$   \\ \hline \hline
$\mathbf{P}_{\textit{predict}}$ & $32 \rightarrow 1024 \rightarrow 512 \rightarrow 256 \rightarrow 22$ \\ \hline
\end{tabular}%
\label{tab:arch}
\caption{Network architecture details for all the sub-networks used in the module.}
\end{wrapfigure}
\subsection{Implementation Details}
All the sub-networks used in proposed module are Multi-Layer perceptrons with ReLU non-linearity. Network architecture for each of the sub-networks are mentioned in Table \ref{tab:arch}. The entire network is trained end to end with the $\mathcal{L}_\textit{E-VAE}$ loss using an ADAM optimizer with a batch size of $512$ and learning rate of $3 \times 10^{-4}$ for all experiments. For the loss coefficient weights, we set $\lambda_1 = \lambda_2 = 1$. We use $N = 3$ rounds of social pooling for Stanford Drone Dataset and $N = 1$ for ETH \& UCY scenes. Using social masking, we perform the forward pass in mini-batches instead of processing all the pedestrians in the scene in a single forward pass (to aboid memory overflow) constraining all the neighbours of a pedestrian to be in the same mini-batch.
\label{sec:metrics}

 \noindent\textbf{Metrics}: For prediction evaluation, we use the Average Displacement Error (ADE) and the Final Displacement Error (FDE) metrics which are commonly used in literature \cite{gupta2018social,alahi2016social,alahi2014socially,li2019conditional}. ADE is the average $\ell_2$ distance between the predictions and the ground truth future and FDE is the $\ell_2$ distance between the predicted and ground truth at the last observed point. Mathematically,  
\begin{equation}
\textit{ADE} = \frac{\sum_{j=t_i + 1}^{t_p + t_f + 1}\Vert \hat{\mathbf{u}}_{j} - \mathbf{u}_{j} \Vert_2}{t_f} \qquad \textit{FDE} = \Vert \hat{\mathbf{u}}_{t_p + t_f + 1} - \mathbf{u}_{t_p + t_f + 1} \Vert_2 
\end{equation}

\begin{table}[t!]
\centering
\resizebox{\textwidth}{!}{%
\begin{tabular}{|c|c|c|c|c|c|c||c|c|c|c|}
\hline
    & SoPhie & S-GAN  & DESIRE  & CF-VAE* & P2TIRL$^\dagger$& SimAug$^\dagger$ & O-S-TT & O-TT & Ours  & PECNet (Ours) \\ \hline 
K   & 20     &     20 &    5    &    20  &   20.  &  20    &    20     & 20      &     5 &           20  \\ \hline
ADE & 16.27  &  27.23 & 19.25   & 12.60  & 12.58  & 10.27  &    10.56     & 10.23      & 12.79 & \textbf{9.96}   \\
FDE & 29.38  &  41.44 & 34.05   & 22.30  & 22.07  & 19.71  &    16.72     & 16.29      & 25.98 & \textbf{15.88}   \\ \hline
\end{tabular}%
}
\vspace{1mm}
\caption{Comparison of our method against several recently published multi-modal baselines and previous state-of-the-art method (denoted by *) on the Stanford Drone Dataset \cite{robicquet2016learning}. `-S' \& `-TT' represents ablations of our method without social pooling \& truncation trick. We report results for in pixels for both $K=5$ \& $20$ and for several other $K$ in Figure \ref{fig:num_samples}. $\dagger$ denotes concurrent work. Lower is better. }
\label{tab:sdd}
\vspace{-5mm}
\end{table}
\noindent where $\mathbf{u}_{j}$, $\hat{\mathbf{u}_{j}}$ are the ground truth and our estimated position of the pedestrian at future time step $j$ respectively.   

\noindent\textbf{Baselines}: We compare our PECNet against several published baselines including previous state-of-the-art methods briefly described below. 
\begin{itemize}
    \item Social GAN (S-GAN) \cite{gupta2018social}: Gupta \etal propose a multi-modal human trajectory prediction GAN trained with a variety loss to encourage diversity.
    \item SoPhie \cite{sadeghian2019sophie}: Sadeghian \etal propose a GAN employing attention on social and physical constraints from the scene to produce human-like motion.
    \item CGNS \cite{li2019conditional}: Li \etal posit a Conditional Generative Neural System (CGNS) that uses conditional latent space learning with variational divergence minimization to learn feasible regions to produce trajectories. They also established the previous state-of-the-art result on the ETH/UCY datasets.
    \item DESIRE \cite{lee2017desire}: Lee \etal propose an Inverse optimal control based trajectory planning method that uses a refinement structure for predicting trajectories. 
    \item CF-VAE \cite{bhattacharyya2019conditional}: Recently, a conditional normalizing flow based VAE proposed by Bhattacharyya \etal pushes the state-of-the-art on SDD further. Notably, their method also does not also rely on the RGB scene image. 
    \item P2TIRL \cite{deo2020trajectory}: A concurrent work by Deo \etal proposes a method for trajectory forecasting  using a grid based policy learned with maximum entropy inverse reinforcement learning policy. They closely tie with the previous state-of-the-art \cite{bhattacharyya2019conditional} in ADE/FDE performance. 
    \item SimAug \cite{liang2020simaug}: More recently, a concurrent work by Liang \etal proposes to use additional 3D multi-view simulation data adversarially, for novel camera view adaptation. \cite{liang2020simaug} improves upon the P2TIRL as well, with performance close to PECNet's base model. However our best model (with pooling and truncation) still achieves a better ADE/FDE performance.        
    \item Ours-TT: This represents an ablation of our method without using the truncation trick. In other words, we set $\sigma_T$ to be identically $1$ for all $K$ settings. Truncation trick ablations with different $K$ are shown in Fig \ref{fig:num_samples} \& Table \ref{tab:sdd}.
    \item Ours-S-TT: This represents an ablation of our method without using both the social pooling module and the truncation trick \ie the base PECNet. We set $\sigma_T = 1$ and $N = 0$ for the number of rounds of social pooling and directly transmit the representations to $\mathbf{P}_\textit{future}$, the prediction sub-network. 
\end{itemize}

\subsection{Quantitative Results}
In this section, we compare and discuss our method's performance against above mentioned baselines on the ADE \& FDE metrics. 
\\

\noindent\textbf{Stanford Drone Dataset}: Table \ref{tab:sdd} shows the results of our proposed method against the previous baselines \& state-of-the-art methods. Our proposed method achieves a superior performance compared to the previous state-of-the-art \cite{bhattacharyya2019conditional,deo2020trajectory} on both ADE \& FDE metrics by a significant margin of \textbf{20.9\%}. Even without using the proposed social pooling module \& truncation trick (OUR-S-TT), we achieve a very good performance ($10.56$ ADE), underlining the importance of future endpoint conditioning in trajectory prediction.
\setlength{\columnsep}{7pt}
\setlength{\intextsep}{0pt}
\begin{wrapfigure}{r}{0.5\textwidth}
  \begin{center}
  \includegraphics[width=0.55\textwidth]{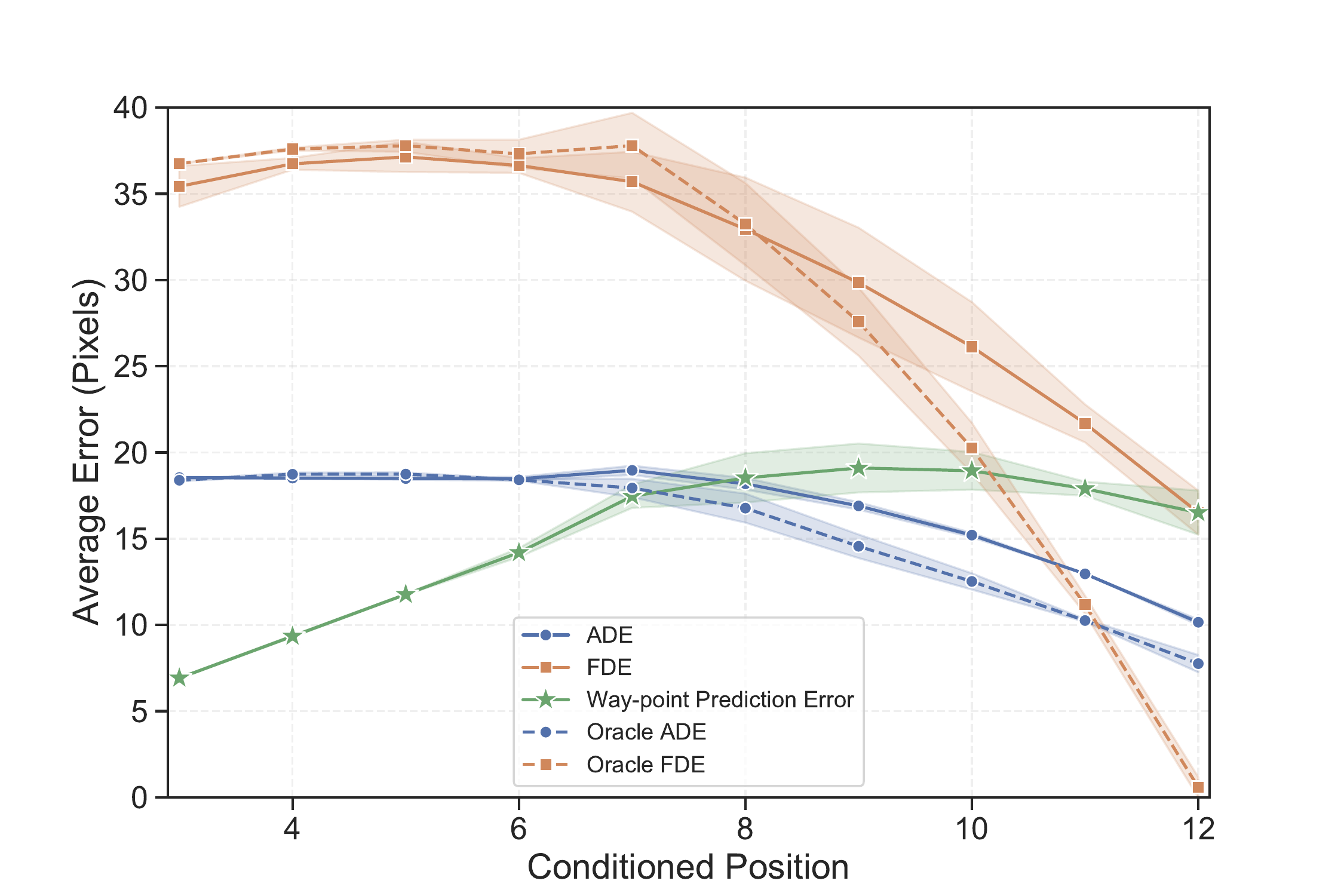}
\end{center}
\setlength{\intextsep}{0pt}
\caption{\small {\bf Conditioned Way-point positions \& Oracles}: We evaluate the performance of the proposed method against the choice of future conditioning position on ADE \& FDE metrics. Further, we evaluate the performance of a destination oracle version of the model that receives perfect information on conditioned position for predicting rest of the trajectory.}
\label{fig:waypoints}
\end{wrapfigure}
As observed by the difference in performance between Ours-S-TT and Our-TT, the social pooling module also plays a crucial role, boosting performance by $0.33$ ADE ($\sim2.1$\%). Note that, while both P2TIRL \cite{deo2020trajectory} \& SimAug \cite{liang2020simaug} are concurrent works, we compare with their methods' performance as well in Table \ref{tab:sdd} for experimental comprehensiveness. All reported results averaged for $100$ trials.
\\

\noindent \textbf{ETH/UCY}: Table \ref{tab:eth-ucy} shows the results for evaluation of our proposed method on the ETH/UCY scenes. We follow the leave-one-out evaluation protocol with $K=20$ as in CGNS \cite{li2019conditional}/Social-GAN \cite{gupta2018social}. All reported numbers are \textit{without} the truncation trick. In this setting too, we observe that our method outperforms previously proposed methods, including the previous state-of-the-art \cite{li2019conditional}. We push the state-of-the-art on average by $\sim \textbf{40.8\%}$ with the effect being the most on HOTEL ($74.2\%$) and least on ETH ($12.9\%$). Also, without the social pooling \& truncation trick (OUR-S-TT) the performance is still superior to the state-of-the-art by $34.6\%$, underlining the usefulness of conditioning on the endpoint in PECNet.
\begin{table}[t!]
\centering
\resizebox{\textwidth}{!}{%
\begin{tabular}{|c|c|c|c|c|c|c|c|c||c|c|c|c|}
\hline
& \multicolumn{2}{c|}{S-GAN} & \multicolumn{2}{c|}{SoPhie}     & \multicolumn{2}{c|}{CGNS*}       & \multicolumn{2}{c||}{S-LSTM} & \multicolumn{2}{c|}{Ours - S - TT}     & \multicolumn{2}{c|}{PECNet (Ours)}      \\ \hline
\multicolumn{1}{|l|}{} & \multicolumn{1}{l|}{ADE} & FDE  & \multicolumn{1}{l|}{ADE} & FDE  & \multicolumn{1}{l|}{ADE} & FDE  & \multicolumn{1}{l|}{ADE}  & FDE  & \multicolumn{1}{l|}{\ ADE\ \ } & FDE  & \multicolumn{1}{l|}{\ \ ADE\ \ \ } & FDE \\ \hline
ETH                    & 0.81                     & 1.52 & 0.70                     & 1.43 & 0.62                     & 1.40 & 1.09                      & 2.35 & 0.58  &  0.96                       & \textbf{0.54}                       & \textbf{0.87}  \\ 
HOTEL                  & 0.72                     & 1.61 & 0.76                     & 1.67 & 0.70                     & 0.93 & 0.79                      & 1.76 & 0.19 & 0.34                         & \textbf{0.18}                      & \textbf{0.24} \\ 
UNIV                   & 0.60                     & 1.26 & 0.54                     & 1.24 & 0.48                     & 1.22 & 0.67                      & 1.40 & 0.39 &0.67                          & \textbf{0.35}                          &  \textbf{0.60}   \\ 
ZARA1 & 0.34 & 0.69 & 0.30 & 0.63 & 0.32   & 0.59 & 0.47 & 1.00 & 0.23 & 0.39 &  \textbf{0.22}   & \textbf{0.39} \\ 
ZARA2 & 0.42 & 0.84 & 0.38 & 0.78 & 0.35 & 0.71 & 0.56 & 1.17 & 0.24 & 0.35 &\textbf{0.17} & \textbf{0.30} \\ \hline \hline
AVG & 0.58 & 1.18 & 0.54 & 1.15 & 0.49 & 0.97 & 0.72 & 1.54 & 0.32 & 0.54 & \textbf{0.29}  & \textbf{0.48} \\ \hline
\end{tabular}%
}
\vspace{1mm}
\caption{Quantitative results for various previously published methods and state-of-the-art method (denoted by *) on commonly used trajectory prediction datasets. Both ADE and FDE are reported in metres in world coordinates. `Our-S-TT' represents ablation of our method without social pooling \& truncation trick.}
\label{tab:eth-ucy}
\vspace{-8mm}
\end{table}

\noindent \textbf{Conditioned Way-point positions \& Oracles}: For further evaluation of our model, we condition on future trajectory points other than the last observed point which we refer to as \textit{way-points}. Further, to decouple the errors in inferring the conditioned position from errors in predicting a path to that position, we use a destination (endpoint) oracle. The destination oracle provides ground truth information of the conditioned position to the model, which uses it to predict the rest of the trajectory. All of the models, with and without the destination oracle are trained from scratch for each of the conditioning positions.     

 \noindent Referring to Figure \ref{fig:waypoints}, we observe several interesting and informative trends that support our earlier hypotheses. (A) As a sanity check, we observe that as we condition on positions further into the future, the FDE for both the Oracle model \& the proposed model decrease with a sharp trend after the $7$th future position. This is expected since points further into the future provide more information for the final observed point. (B) The ADE error curves for both the oracle and the proposed model have the same decreasing trend albeit with a gentler slope than FDE because the error in predicting the other points (particularly the noisy points in the middle of the trajectory) decreases the gradient. 
 \setlength{\columnsep}{10pt}
\setlength{\intextsep}{0pt}
\begin{wrapfigure}{l}{0.5\textwidth}
  \begin{center}
  \includegraphics[width=0.55\textwidth]{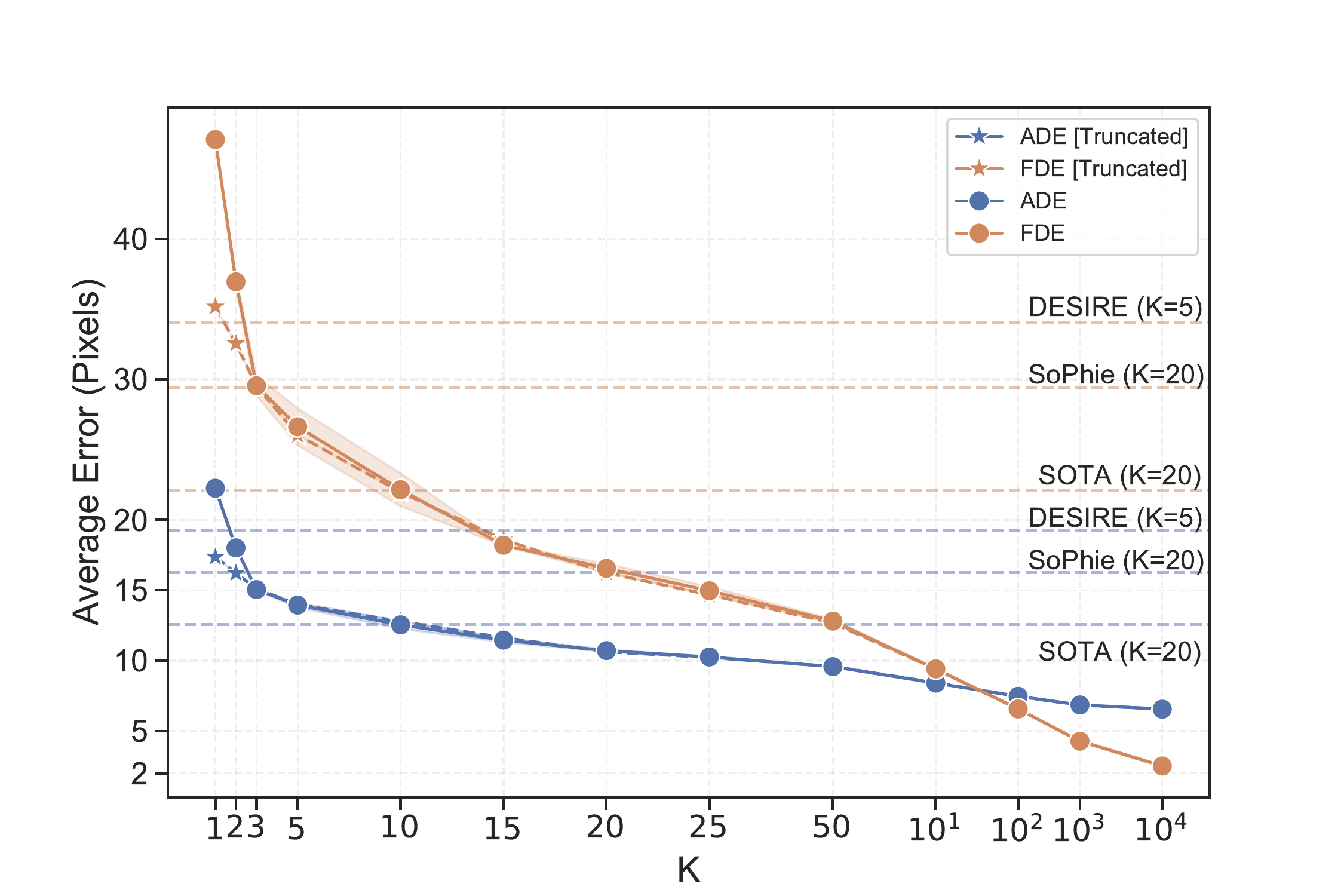}
\end{center}
\setlength{\intextsep}{0pt}
\caption{\small {\bf Performance across $K$}: ADE \& FDE performance of our method against number of samples used for evaluation. Several previous baselines are mentioned as well with their number of samples used. Our method significantly outperforms the state-of-the-art reaching their performance with much lesser number of samples \& performing much better with same number of samples as theirs ($K = 20$).}
\label{fig:num_samples}
\end{wrapfigure}
(C) Interestingly, our model's ADE and FDE is not significantly different from that of the Oracle model for points close in the future and the error in the two models are approximately the same until about the $7$th future position. This suggests that till around the middle of the future, the conditioned way-points do not hold significant predictive power on the endpoint and hence using our noisy guesses vs. the oracle's ground truth for their position does not make a difference. 

\noindent \textbf{Way-point Prediction Error}: The way-point position error is the $\ell_2$ distance between the prediction of location of the conditioned position and its ground truth location (in the future). Referring to Figure \ref{fig:waypoints}, we observe an interesting trend in the way-point error as we condition on points further into the future. The way-point prediction error increases at the start which is expected since points further into the future have a higher variance. However, after around the middle ($7$th point) the error plateaus and then even slightly decreases. This lends support to our hypothesis that pedestrians, having predilection towards their destination, exert their will towards it. Hence, \textit{predicting the last observed way-point allows for lower prediction error} than way-points in the middle! This in a nutshell, confirms the motivation of this work.  
\\

\noindent \textbf{Effect of Number of samples (K)}: All the previous works use $K=20$ samples (except DESIRE which uses $K=5$) to evaluate the multi-modal predictions for metrics ADE \& FDE. Referring to Figure \ref{fig:num_samples}, we see the expected decreasing trend in ADE \& FDE with time as $K$ increases. Further, we observe that our proposed method achieves the same error as the previous works with much smaller $K$. Previous state-of-the-art achieves $12.58$ \cite{deo2020trajectory} ADE using $K=20$ samples which is matched by PECNet at half the number of samples, $K = 10$. This further lends support to our hypothesis that conditioning on the inferred way-point significantly reduces the modeling complexity for multi-modal trajectory forecasting, providing a better estimate of the ground truth.
\begin{figure}[b!]
\vspace{-4mm}
    \centering
    \includegraphics[width=\textwidth]{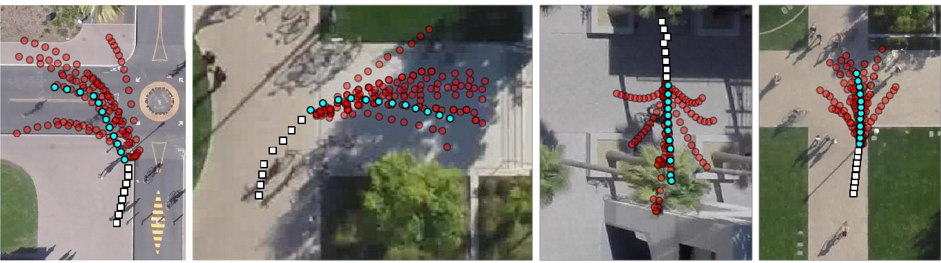}
    \includegraphics[width=\textwidth]{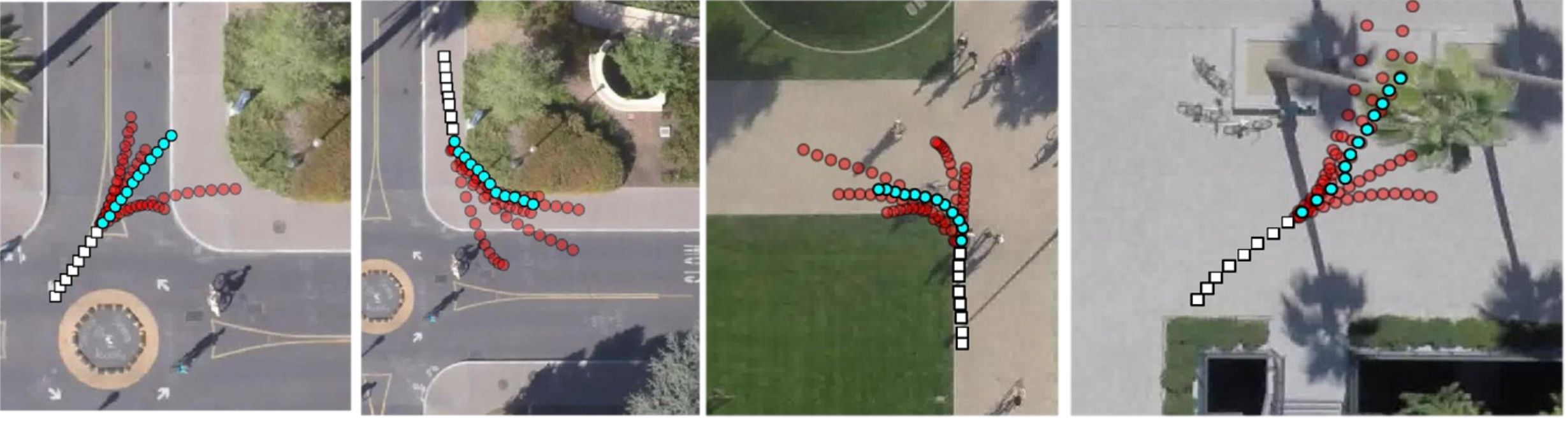}
    \caption{\small{\textbf{Visualizing Multimodality}}: We show visualizations for some multi-modal and diverse predictions produced by PECNet. White represents the past $3.2$ seconds while red \& cyan represents predicted \& ground truth future respectively over next $4.8$ seconds. Predictions capture a wide-range of plausible trajectory behaviours while discarding improbable ones like, endpoints opposite to pedestrian's direction of motion.} 
    \label{fig: multimodality}
\end{figure}

\noindent Lastly, as $K$ grows large ($K \rightarrow \infty$) we observe that the FDE slowly gets closer to $0$ with more number of samples, as the ground truth $\mathcal{G}_c$ is eventually found. However, the ADE error is still large ($6.49$) because of the errors in the rest of the predicted trajectory. This is in accordance with the observed ADE ($8.24$) for the oracle conditioned on the last observed point (\ie $0$ FDE error) in Fig. \ref{fig:waypoints}. 
\\

\noindent \textbf{Design choice for VAE}: We also evaluate our design choice of using the inferred future way-points $\hat{\mathcal{G}_c}$ for training subsequent modeules (social pooling \& prediction) instead of using the ground truth ${\mathcal{G}_c}$. As mentioned in Section \ref{sec:method_joint_traj}, this is also a valid choice for training PECNet end to end. Empirically, we find that such a design achieves $10.87$ ADE and $17.03$ FDE. This is worse ($\sim 8.8\%$) than using $\hat{\mathcal{G}_c}$ which motivates our design choice for using $\hat{\mathcal{G}_c}$ (Section \ref{sec:method_joint_traj}).  

\noindent \textbf{Truncation Trick}: Fig. \ref{fig:num_samples} shows the improvements from the truncation trick for an empirically chosen hyperparameter $c \approx 1.2$. As expected, small values of $K$ gain the most from truncation, with the performance boosting from 22.85 ADE ($48.8$ FDE) to $17.29$ ADE ($35.12$ FDE) for $K = 1$ ($\sim 24.7\%$).
\vspace{-2.5mm}
\subsection{Qualitative Results}

 In Figure \ref{fig: multimodality}, we present several visualizations of the predictions from PECNet. As shown, PECNet produces  multi-moda diverse predictions conditioning on inferred endpoints. In Figure \ref{anim:one}, we present animations of several socially compliant predictions. The visualizations show that along with producing state-of-the-art results, PECNet can also perform rich multi-modal multi-agent forecasting.
 \begin{figure}[t!]
    \animategraphics[loop,autoplay,width=0.415\textwidth]{8}{appendix_1/}{1}{91}
    \animategraphics[loop,autoplay,width=0.585\textwidth]{8}{appendix_2/}{1}{54}
\caption{\small{\textbf{Diverse Mutltimodal \& Social Interactions}}: \small{Visualizations denoting multiple socially compliant trajectories predicted with PECNet. Left pane shows future trajectories for $9.6$ seconds predicted in a recurrent input fashion. Right pane shows the predicted trajectories for future $4.8$ seconds at an intersection. Solid circles represent the past input \& stars represent the future ground truth. Predicted multi-modal trajectories are shown as translucent circles jointly for all present pedestrians. Animation is best viewed in \textit{Adobe Acrobat Reader}. More video visualizations available at project homepage: \href{https://karttikeya.github.io/publication/htf/}{https://karttikeya.github.io/publication/htf/}}} 
\label{anim:one}
\vspace{-4mm}
\end{figure}
\vspace{-4mm}
\section{Conclusion}
\vspace{-2mm}
In this work we present PECNet, a pedestrian endpoint conditioned trajectory prediction network. We show that PECNet predicts rich \& diverse multi-modal socially compliant trajectories across a variety of scenes. Further, we perform extensive ablations on our design choices such as endpoint conditioning position, number of samples \& choice of training signal to pinpoint achieved performance gains. We also introduce a ``truncation trick" for trajectory prediction, a simple method for adjusting diversity for performance in trajectory prediction without retraining. Finally, we benchmark PECNet across multiple datasets including Stanford Drone Dataset \cite{robicquet2016learning}, ETH \cite{pellegrini2010improving} \& UCY \cite{lerner2007crowds} in all of which PECNet achieved state-of-the-art performance.
\clearpage
\bibliography{egbib}
\end{document}